\documentclass[runningheads]{llncs}
\usepackage{graphicx}

\usepackage{hyperref}
\hypersetup{
	colorlinks   = true, 
}
\usepackage{color}

\usepackage[width=122mm,left=12mm,paperwidth=146mm,height=193mm,top=12mm,paperheight=217mm]{geometry}

\def\etal{\emph{et al}. }
\usepackage[linesnumbered,ruled]{algorithm2e}

\begin{document}

\title{Revisiting Distillation and Incremental Classifier Learning} 

\titlerunning{Revisiting iCaRL}

\authorrunning{K. Javed, F. Shafait}

\author{Khurram Javed$^2$, Faisal Shafait$^{1,2}$}


\institute{$^1$Deep Learning Laboratory, National Center of Artificial Intelligence, \\ Islamabad \\ $^2$School of Electrical Engineering and Computer Science,\\
	National University of Sciences and Technology\\
	\email{\{14besekjaved, faisal.shafait\}@seecs.edu.pk}\\
}

\maketitle


\begin{abstract}
	One of the key differences between the learning mechanism of humans and Artificial Neural Networks (ANNs) is the ability of humans to learn one task at a time. ANNs, on the other hand, can only learn multiple tasks simultaneously. Any attempts at learning new tasks incrementally cause them to completely forget about previous tasks. This lack of ability to learn incrementally, called Catastrophic Forgetting, is considered a major hurdle in building a true AI system.
	\par In this paper, our goal is to isolate the truly effective existing ideas for incremental learning from those that only work under certain conditions. To this end, we first thoroughly analyze the current state of the art (iCaRL) method for incremental learning and demonstrate that the good performance of the system is not because of the reasons presented in the existing literature. We conclude that the  success of iCaRL is primarily due to knowledge distillation and recognize a key limitation of knowledge distillation, i.e, it often leads to bias in classifiers. Finally, we propose a dynamic threshold moving algorithm that is able to successfully remove this bias. We demonstrate the effectiveness of our algorithm on CIFAR100 and MNIST datasets showing near-optimal results. Our implementation is available at : \url{https://github.com/Khurramjaved96/incremental-learning}.
\end{abstract}

\section{Introduction}
\label{sec:intro}
To understand incremental learning, let's look at a simple everyday example. Suppose that you are taking a walk in a garden and you come across a new kind of flower. You have never seen such a flower before and are intrigued by it, so you look up all the information regarding that flower, and learn everything you possibly can about it. You then continue your walk and come across a red rose; would you be able to recognize the rose? Assuming you have seen a rose in the past, the answer is a resounding yes. In fact, the question seems completely unrelated to the task that you learned recently. However, if an Artificial Neural Network (ANN) was asked to do the same thing, it won't be able to answer the question. Even a network trained to recognize roses with an accuracy of 100\% would fail to answer the question if it was not provided samples of roses at the exact moment it was learning about the new flower. This phenomenon is known as Catastrophic Forgetting and highlights a sharp contrast between the way humans and neural networks learn; humans are able to attain new knowledge without forgetting previously stored information. Artificial Neural Networks, trained using a variation of Gradient descent, on the other hand, must be provided with data of all previously learned tasks whenever they are learning something new (It should be noted that humans also have to revise old tasks eventually to be able to retain knowledge over long periods of times and that the new knowledge can in fact interfere with older knowledge \cite{wimber2015retrieval}. However, the problem is not nearly as pronounced and debilitating in humans as it is in ANNs.). The goal of incremental learning is to bridge this gap between humans and ANNs. 

The importance of incorporating incremental learning in ANNs is self evident; not only will it address a key limitation of ANNs and a fundamental AI research problem, but also provide countless practical benefits such as deploying ever evolving machine learning systems that can dynamically learn new tasks over time, or developing classifiers that can handle an ever changing set of classes (For example inventory of a store).

\paragraph{} In this paper, instead of tackling the general problem of incremental learning of multiple tasks as described above, we limit our discussion to incremental classifier learning. We believe this is justified because incremental classifier learning presents most of the same challenges as the general incremental learning problem, and is at the same time easier to tackle given the current state of ANNs. Note that we are not the first one to limit ourselves to incremental classifier learning and as we shall see later in the paper, some of the most popular work on incremental learning made the same simplifying assumption.

\subsection{Our Contributions:} We make three main contributions in this work. First, we analyze the existing state of the art for incremental classifier learning, iCaRL \cite{rebuffi2017icarl} and make some insightful observations about the proposed approach. We then propose a novel Dynamic Threshold Moving Algorithm to address a well known issue of a popular knowledge transfer and preservation technique called Knowledge Distillation. Finally, we present a simple solution to address the major issue of lack of reproducibility in scientific literature. 
\begin{enumerate}
	\item 	\textbf{Analysis of iCaRL:} We thoroughly analyze the current state of the art incremental learning approach proposed by S. Rebuff \etal \cite{rebuffi2017icarl} and show that some of the improvements resulting from this approach are not because of the reasons presented by the authors. More specifically, we show that NEM (Nearest Exemplar Mean) classifier is only effective because the classifier learned through the training procedure is biased, and by either implementing threshold moving or using higher temperature distillation, it is possible to remove this bias. As a result, NEM classifier is not a necessary requirement for an incremental classifier. The author also proposed an exemplar set selection algorithm, called herding, well suited to approximate NCM (Nearest Class Mean) classifier. We failed to reproduce the effectiveness of herding in our experiments. In fact, herding did not perform any better than random instance selection. W. Yue \etal \cite{neu} also tried to independently reproduce the results of herding but failed. 
	\\
	\item \textbf{Dynamic Threshold Moving Algorithm:} We propose an algorithm for computing a scale vector that can be used to fix the bias of a classifier trained using distillation loss. The problem of bias resulting from distillation was first noticed by G. Hinton \etal \cite{hinton2015distilling} in their original work on knowledge distillation. However, in their work, they only showed the existence of a vector $\mathcal S$ that can be used to fix the bias. They did not provide a method for computing the said vector. Using our algorithm, on the other hand, it is possible to compute the said vector at no additional cost. 
	\\
	\item \textbf{Framework for Future Work: }  We open-source an implementation of class incremental learning that can be used to quickly compare existing methodologies on multiple datasets. We develop our framework keeping in mind ease of extensibility to newer datasets and methods, and propose a simple protocol to facilitate quick reproducibility of results. We hope that future researchers would follow a similar protocol to complement the already positive trend of code, data, and paper sharing for quick reproducibility in the machine learning community.
\end{enumerate}

\paragraph{}	
We are confident that our work clarifies some of the uncertainties regarding incremental classifier learning strategies, and would act as a stepping stone for future research in incremental learning. We're also hopeful that our dynamic threshold moving algorithm will find other use-cases than that of training a classifier with distillation. One such potential use-case is to use dynamic threshold for removing bias when transferring knowledge to a student model from a larger, deeper teacher model. 

\section{Related Work}
The problem of catastrophic forgetting was identified as early as 1989 by McCloskey \etal \cite{mccloskey1989catastrophic}. This led to preliminary work on incremental representation learning in the late 90s \cite{old1} \cite{old2} \cite{old3}. Later in 2013, Goodfellow \etal \cite{cur1} extended the work to include current, deeper and more sophistical ANNs by presenting a thorough empirical analysis of Catastrophic Forgetting.  

Methods for incremental classifier learning were also proposed quite early. For example, T. Mensink \etal \cite{mensink2012metric} demonstrated that by fixing the representation and using NCM Classifier, it is possible to add new classes at zero additional cost. C.H Lampert \etal \cite{zero} proposed a zero shot learning system that was able to classify new classes at no cost. However, in both of these methods, the feature extraction pipeline was not adapted to the new data, and only the classification algorithm was changed to incorporate new classes. This limited the performance of the system as feature extractor trained on one set of classes may not generalize to a newer set.  

The remaining recent Incremental learning techniques can be categorized into (1) Rehearsal based and (2) Knowledge preserving incremental learning. 
\subsection{Rehearsal Based Incremental Learning}
Rehearsal based incremental learning systems have a pipeline similar to joint training. They store the distribution of the data of the previously learned tasks in some form, and use samples from the stored distribution of the previous tasks to avoid catastrophic forgetting.

The most notable example of the rehearsal based system is iCaRL \cite{rebuffi2017icarl}. ICaRL stores a total of $K$ number of exemplars from previously seen classes and uses distillation loss \cite{hinton2015distilling} in a way similar to Learning without Forgetting \cite{li2017learning} to retain the knowledge of previous classes. L. David and R. Marc \cite{fb} recently proposed a rehearsal based method that allows for positive backward transfer (i.e, the performance of the system improves on older tasks as it learns new tasks). However, they assumed that task descriptors are available at test time. This makes their problem statement only a very special case of ours. 

GANs have recently become popular for storing distribution of the data. R. Venkatesan \etal \cite{gan} and W. Yue \etal \cite{neu} proposed incremental learning using GANs. Instead of storing $K$ exemplars similar to iCaRL, they propose training a GAN that can generate data for the older tasks. They then use this GAN to generate images of older classes for rehearsal. However, because of the current limitations of GANs on complex datasets, purely GAN based approaches are not competitive yet.

\subsection{Knowledge Preserving Incremental Learning}
The second common category of techniques for incremental learning tries to retain the knowledge of the network when learning a new class. One recent approach, proposed by T. Rannen \etal  \cite{rannen2017encoder} uses auto-encoders to preserve the knowledge useful for previous tasks. In this approach, the authors propose learning an under-complete auto-encoder that learns a low dimensional manifold for each task. When learning new tasks, the model tries to preserve the projection of data in this low dimensional manifold, while allowing the feature map to freely change in other dimensions. Oquab \etal \cite{oquab2014learning} proposed that it is possible to minimize forgetting by freezing the earlier and mid-level layers of models. However, this limits the representation learning capability of the system for newer classes. Some researchers also proposed approaches that kept track of important weights for older tasks, and made it harder for the model to update those weights when learning new tasks \cite{zenke2017improved} \cite{kirkpatrick2017overcoming}. Finally, distillation loss \cite{hinton2015distilling}, inspired by the work of Caruana \etal \cite{buciluǎ2006model}, can be used for retaining older knowledge. For example, Zhizhong \etal \cite{li2017learning} showed that by computing distillation using a copy of an earlier model, it is possible to retain knowledge of older tasks. Recently, R. Kemker and C. Kanan \cite{kemker2017fearnet} proposed a methodology inspired by the human brain to solve incremental learning. However in their work, they use a pre-trained imagenet model as feature extractor and as a result, do not tackle incremental representation learning. Finally, A. Rusu \etal \cite{grow2} and T. Xiao \cite{grow1} proposed networks that grew as new classes were added to avoid changing weights important for older tasks. Their method is not memory bounded, however. 
\section{Overview of iCaRL} 
\begin{figure}
	\label{herdingalgo}
	\begin{algorithm}[H]
		\label{herding}
		{\bf Input:} Trained Model $M$, $C_i^j \in$ Images of class $i$ , Size $k$\;
		{\bf Output:} Set containing $k$ instances of class $C_i$\;
		{ $\forall C_i^j \in C_i$, use $M$ to get the feature map $F_i^j$}\;
		{ Let $S$ be a null set}\;
		{ Compute the mean of  all $F_i^j$. Let this be $F_i^{mean}$}\;
		{ Select $F_i^j$ and add it in $S$ such that mean of selected set is closest to $F_i^{mean}$}\;
		{ If $|S|<k$, repeat step $5$. Else, return $S$. }
		
		\caption{Herding Algorithm for Instance Selection}
	\end{algorithm}
	
\end{figure}
In iCaRL, S. Rebuffi \etal \cite{rebuffi2017icarl} define an incremental classifier to satisfy two properties; First, at any time, the system should be able to give a reasonable classification performance for the classes seen so far. Secondly, the memory and computation requirement of the system should stay bounded. To specify the memory bound of the system, they propose a hyper-parameter $K$ called the memory-budget of the classifier. This budget specifies how many instances of the old data, at max, the system is allowed to store at any given time. 

In their implementation, they propose storing $\frac{K}{m}$ instances of each class where $m$ is the number of classes. They call these instances the exemplar set, and use an algorithm called herding to construct the set. During an increment, they use both the new data and the stored exemplars to compute two kinds of losses. One is the standard classification loss computed using the ground truth of the images, whereas the other is a \textit{variant} of the distillation loss proposed by Hinton \etal \cite{hinton2015distilling}. In their implementation of the distillation loss, they don't use the final softmax layer with softened distribution as originally proposed by G. Hinton. Instead, they use sigmoid activation for converting final values to probabilities.

To compute the targets for the distillation loss, they propose making a copy of the classifier just before an increment. Finally, they use the exemplar set to approximate the mean embedding of each class in the feature space, and use this approximate mean class embedding for classification at runtime.

\subsection{Contributions of iCaRL}
The authors claim that their systems work well because of three main contributions. 
\begin{enumerate}
	\item  They use Algorithm \ref{herding} to construct an exemplar set that best approximates the real class means in the feature space. 
	\\ 
	\item  They propose a new classification algorithm, called Neared Exemplar Mean (NEM) classifier, for classification. NEM is similar to the well known Nearest Class Mean (NCM)  classifier, except that instead of using true class means for classification, it uses the mean of the exemplar set. The authors claim that NEM classifier gives better results compared to the naive classifier learned through back-propagation during training. 
	\\ 
	\item They use a variation of distillation loss \cite{hinton2015distilling} to transfer knowledge from an older network to the newer, incremented network. However unlike Learning without Forgetting \cite{li2017learning}, they distill knowledge without softening the target probability distribution from the teacher model.  
\end{enumerate}

In the following Section \ref{imple}, we go over the implementation rationale and details of the iCaRL system, and in Section \ref{analysis}, we analyze the first wo contributions made by iCaRL in detail.
\section{Implementation Details}
\label{imple}
We re-implemented the iCaRL paper in PyTorch \cite{pytorch}. To validate our implementation, we compared our results with the ones reported in the iCaRL paper and compared our implementation with the open-sourced implementations of iCaRL. We reimplemented the paper for two main reasons 

First, our goal was to develop an extensible code base over which the research community can easily implement new ideas. To this end, we made sure that the algorithms, models, and datasets modules were decoupled, and that our design choices allowed for addition of new models, algorithms, and datasets without modification of the existing code.

Second, we wanted to introduce a protocol to facilitate quick reproducibility. To achieve this, our implementation automatically generates a meta-data file corresponding to every experiment. This file contains all the parameters that were used to run the experiment, and the hash of the version of the git repository used for the experiment. This means that the minuscule meta-data file contains all the information required to reproduce the results, and by simply sharing the meta-data file along with the paper, the authors would allow others to quickly reproduce their results. 
\subsection{Hyper-parameters Selection}
For our experiments, we used a variant of Resnet32 \cite{resnet} modified to work on CIFAR dataset \cite{cifar} (Similar to iCaRL). Instead of modifying the model to work on MNIST \cite{lecun1998mnist}, we rescaled MNIST images to $32 \times 32$ to work on the same model. All the experiments on CIFAR100 are run for 70 epochs per increment with an initial learning rate of 2.0 and a momentum of 0.9. The learning rate is reduced by a factor of five at epoch 45, 60, and 68. For MNIST, we only run experiments for 10 epochs with an initial learning rate of 0.1 reduced to 0.02 and 0.004 at epoch no 5 and 8. Memory budget of 2,000 is used unless mentioned otherwise.

\section{Analysis of iCaRL} 
\label{analysis}
As discussed above, iCaRL makes three contributions. We discuss and analyze the first two contributions in Section. \ref{exmp} and \ref{nemc} respectively.
\subsection{Exemplar Selection using Herding}
\label{exmp}
\begin{figure}
	\centering
	\includegraphics[width=0.65 \linewidth]{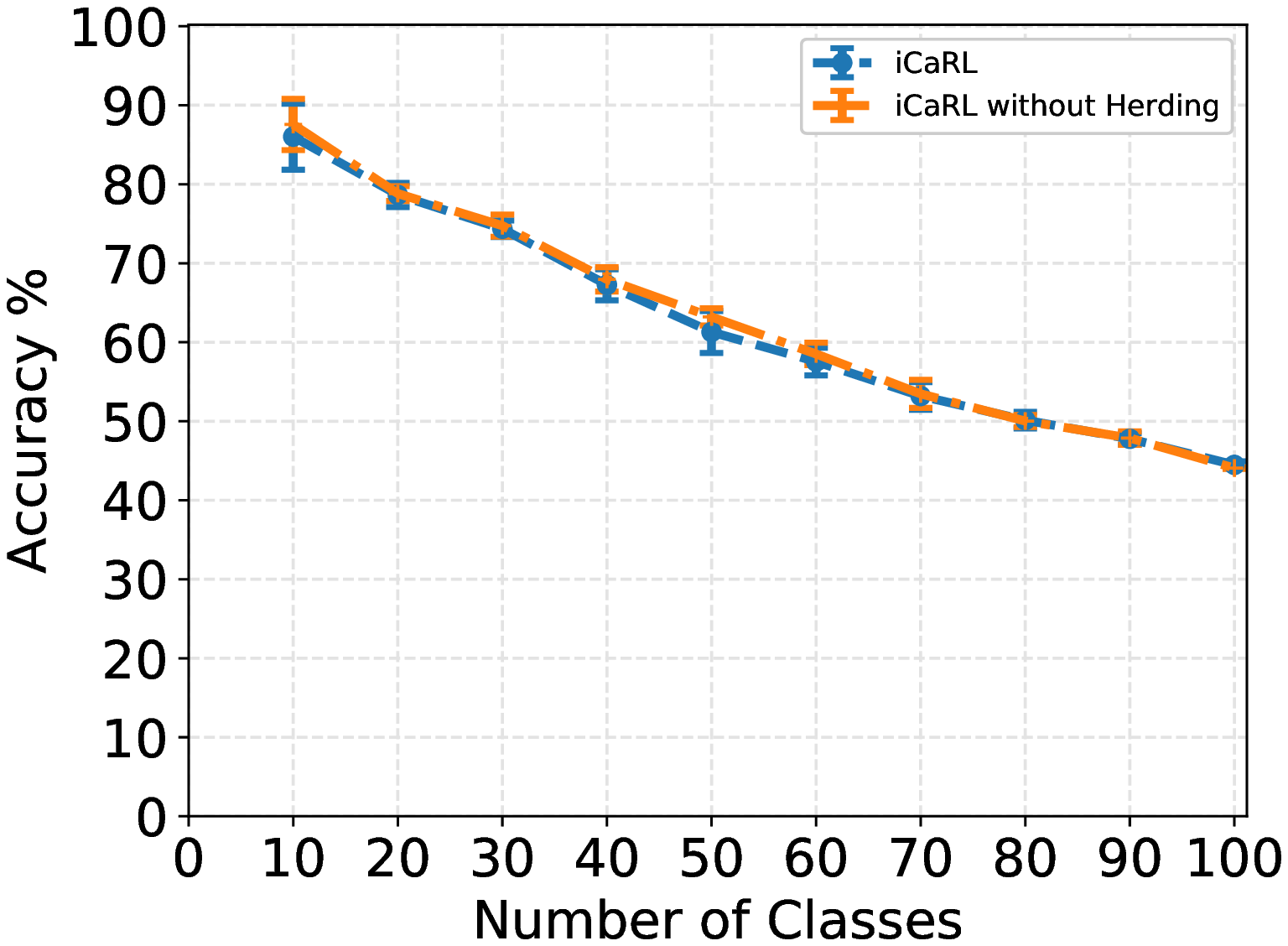}
	\caption{Comparison of iCaRL results with and without herding. There is no significant difference between random exemplar selection and exemplar selection by herding for incremental classifier learning. Here initially the classifier is trained on the first ten classes of CIFAR100, and then ten new classes are added at each increment.}
	\label{fig:herding}
\end{figure}

iCaRL uses the herding algorithm (Algorithm \ref{herding}) for selecting exemplars that approximate the true class mean. We implemented the herding algorithm and tested the performance of the system by toggling herding and keeping all other parameters the same. We discovered that there was no significant difference in the results with or without herding, and random instance selection worked as well as herding. Note that W. Yue \etal \cite{neu} also did a similar experiment and failed to reproduce any improvements resulting from herding. This goes to show that while the authors think that herding helps in choosing an exemplar set that gives an exemplar mean close to the true mean, this is in fact not the case. This makes sense because images chosen to give a good approximation of class mean at increment $i$ will not necessarily give a good approximation at increment $i+1$ because the feature representation of each image would be different after the increment. 
\subsubsection{Experiment Design}
We follow the same experiment design as iCaRL to compare our results. Initially, we train our classifier on the first $p$ randomly chosen classes of CIFAR100 dataset. After training, we construct the exemplar sets by herding or random instance selection for all of the classes that the model has seen so far, and discard the remaining training data. Finally at each increment, we use the complete data of the new $p$ classes, and only the exemplar set for the old classes. 
\subsubsection{Results}
Result of our experiments on CIFAR100 with $p=10$ can be seen in Fig. \ref{fig:herding}. By picking a different order of classes in different runs, we are able to get confidence intervals. Here the error bars correspond to one standard deviation of the multiple runs. Note that both random instance selection and herding give remarkably similar results showing that herding is not necessary for class incremental learning.   

\subsection{Classification using Nearest Exemplar-Set Mean Classifier}
\label{nemc}
The authors of iCaRL claim that an approximate variant of Nearest Class Mean (NCM) classifier that only uses the exemplar set for computing class mean (Let's call this NEM for Nearest Exemplar Mean) is superior to the classifier learned through back-propagation (Let's call this TC for Trained Classifier). To substantiate their claim, they experiment with both NEM and TC and demonstrate that NEM outperforms TC on a number of experiments. For smaller increments (two), NEM is particularly effective over TC in their experiments.

\subsubsection{Hypotheses Development:} Because NEM is just a special case of TC, and in theory, it is possible to learn weights such that TC and NEM are equivalent, we suspected that this large difference between the two might be because the newer classes contribute disproportionately to the loss (Because of the class imbalance; for older classes, we only have the exemplar set). To make matters worse, distillation is also known to introduce bias in training \cite{hinton2015distilling} because the distilled targets may not represent all classes equally. 

Based on these observations, and the hypothesis that NEM performs better because of bias in training, we predict the following: 
\begin{enumerate}
	\item It would be possible to reduce the difference between TC and NEM by using higher temperature (T) when computing distillation. This is because higher values of temperature results in a softer target distribution that represents all classes more fairly.
	\item  It would be possible to improve TC by removing the bias by using some threshold moving strategy. 
\end{enumerate}

We tested both of our predictions experimentally and found conclusive support for our hypothesis. 

\subsubsection{Analysis of  Prediction 1} 
\begin{figure}
	\centering
	\begin{tabular}{cc}
		\includegraphics[width=6.0cm]{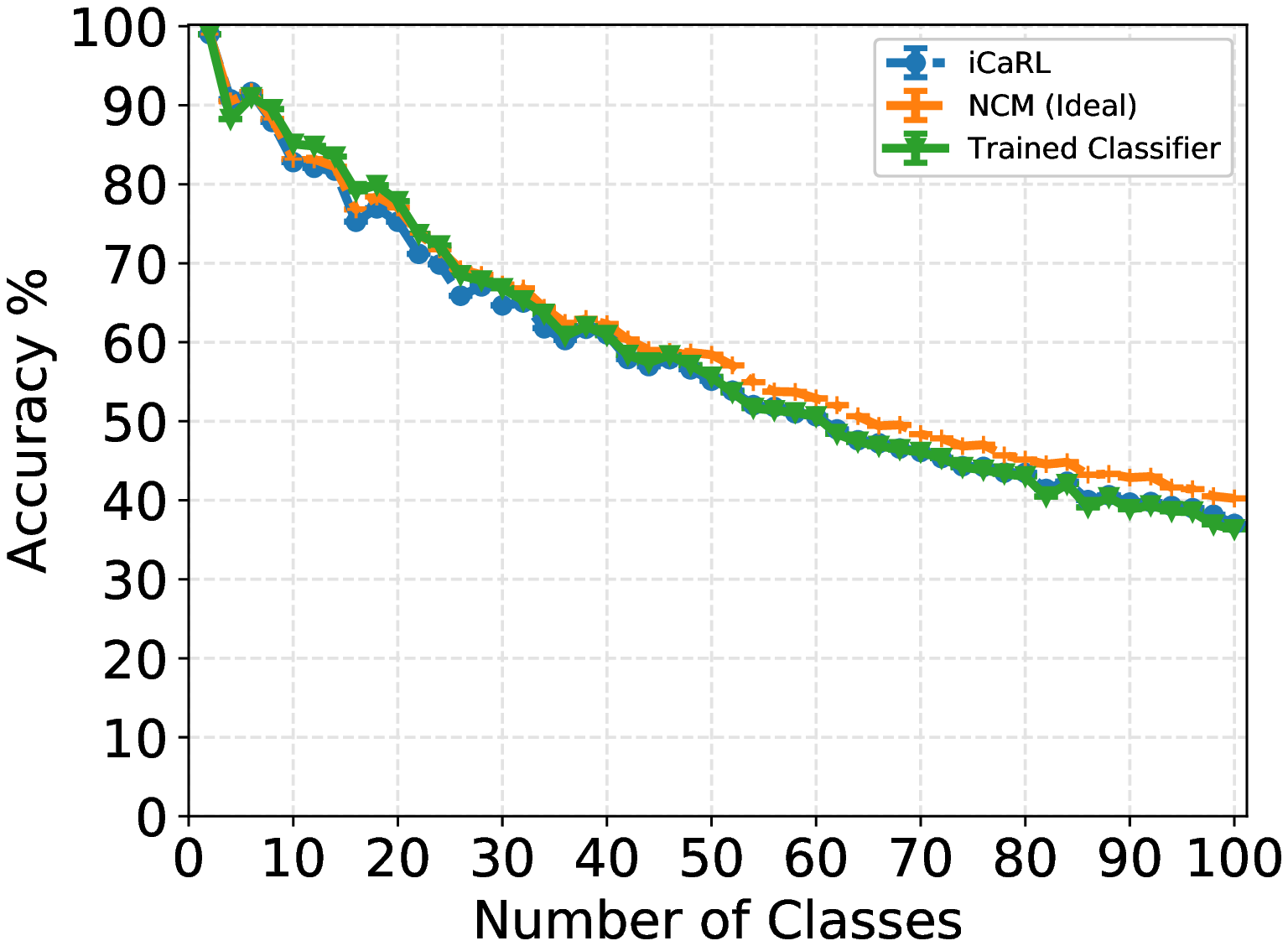}&
		\includegraphics[width=6.0cm]{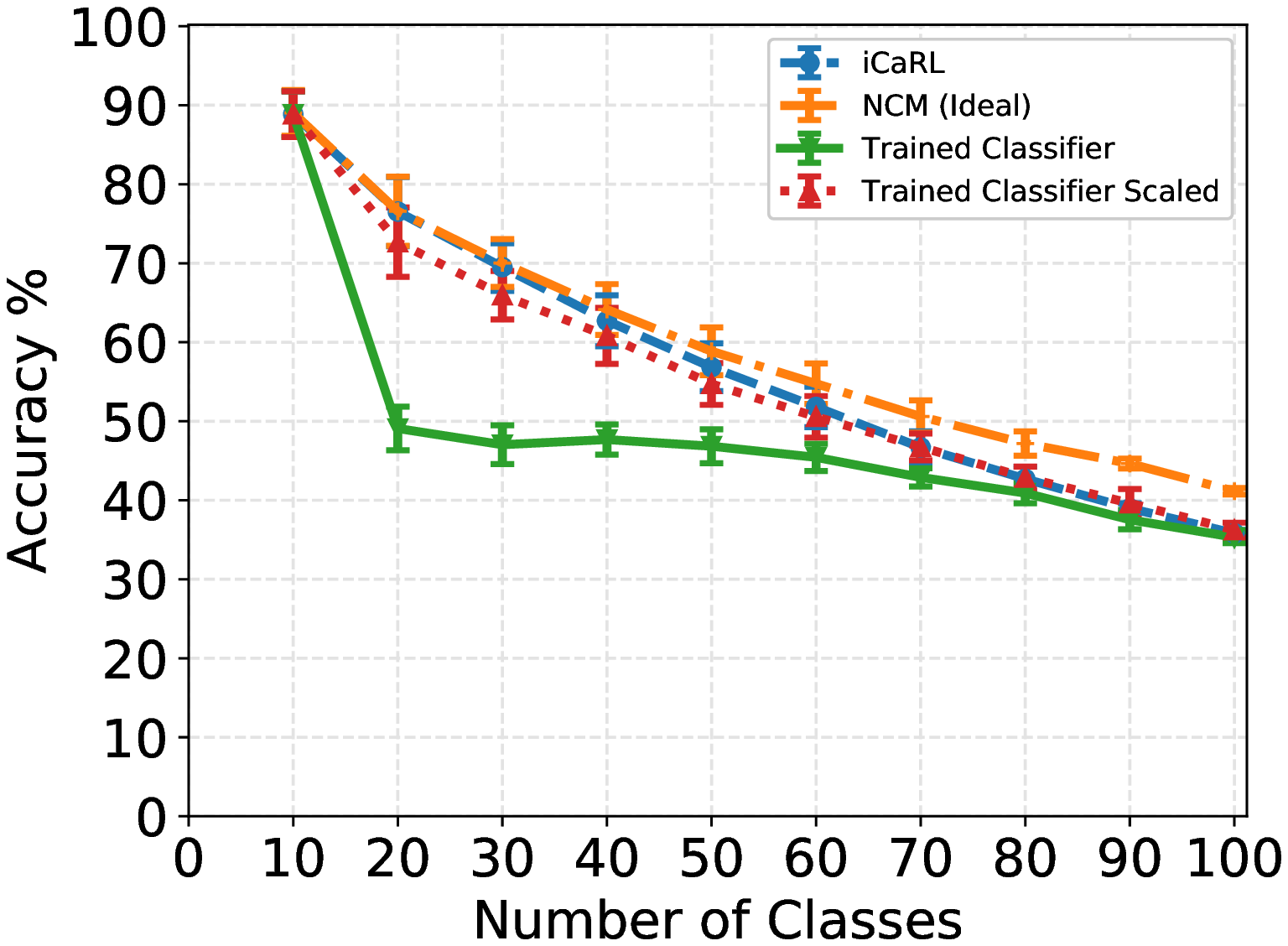}\\
		(a) Bias Removal by Temperature = 3 &(b) Bias Removal by Threshold Moving
	\end{tabular}
	\caption{(a) Comparison of iCaRL NCM with the softmax trained classifier. It is clear that by using temperature values of greater than 1, it is possible to remove the bias from the classifier and get results similar to NEM (iCaRL). (b) Effect of threshold moving (Scaling) using $\mathcal S$ on CIFAR100. We used memory-budget of 500 in this experiment to highlight the bias. Without scaling, iCaRL's NEM indeed does perform better. }
	\label{fig:nmcvstc}
\end{figure}

First, we train an incremental classifier with temperature of 3 and an increment size of 2. Note that as per the authors of iCaRL, this is the worst case for TC and without the higher temperature, NEM outperforms TC by a large margin. With T=3, however, we discovered that there was no difference between TC and NEM as shown in Fig.~\ref{fig:nmcvstc} (a) (The performance was stable across a range of values of T from 3 - 10 in our experiments). In the original implementation of iCaRL without the temperature parameter, TC performed significantly worse as shown in Table.~\ref{tab:t1}.

\subsubsection{Analysis of  Prediction 2} Secondly, we ran an experiment with temperature~=~1 but this time, we scaled the predictions of the model by a vector $\mathcal S$ to remove the bias of the classifier (We present the algorithm for computing $\mathcal S$ in Section 6). Without the scaling, NEM does indeed perform better as claimed by iCaRL. However with scaling, the difference between the two is insignificant. The results of the experiment are shown in Fig. \ref{fig:nmcvstc} (b).

\subsubsection{Conclusion}
From the two experiments, it is evident that while NEM does indeed perform better than TC in some cases, the difference is because of the learned bias of TC and not because NEM is inherently better for class incremental learning. This is an important distinction because in the original paper, the authors attributed the better performance of NEM to the fact that NEM classifier is not decoupled from the data representation. We show that this is not the case, and NEM performs better simply because it is more robust to the bias present at training time. 

\begin{table}
	\caption{By using softmax classifier with high temperature (T=3), the trained classifier in our implementation performs as well as NEM. Note that we only show the results of the worst case scenario for us (i.e, the case when the difference between iCaRL and Trained Classifier was maximum). }
	\begin{center}
		
		\begin{tabular}{|c|c|c|}
			\hline
			Version & iCaRL (NEM) & Trained Classifier \\
			\hline\hline
			iCaRL Implementation & 57.0 & 36.6 \\
			Our Implementation with T=3 & 57.80 &\textbf{ 58.21} \\
			\hline
		\end{tabular}
	\end{center}
	\label{tab:t1}
	
\end{table}
\section{Dynamic Threshold Moving}
In this section, we present a simple yet effective method for computing a vector for threshold moving to overcome the bias learned as a result of distillation. We first give an overview of knowledge distillation, and why distillation loss often leads to this bias. 
\subsection{Knowledge Distillation in Neural Networks} 
Knowledge distillation is a highly popular technique introduced by G. Hinton \etal \cite{hinton2015distilling}  to train small networks with generalization performance comparable to a large ensemble of models. The authors showed that it is possible to achieve significantly better generalization performance using small neural network if the small network was optimized to approximate a pre-trained large ensemble of models on the same task. More concretely, given a large ensemble of models denoted by $F_{ens}(X)$ trained on $x_i, y_i$, where $x_i$ and $y_i$, and a smaller neural network $F_{small}(X)$, it is better to train the smaller network on $(x_i, F_{ens}^T(x_i))$ instead of on the original data $(x_i, y_i)$. Here parameter $T$ denotes that the factor by which the pre-softmax output of the model is divided by before applying the final 'softmax'. In practice, it is better to jointly optimize to approximate the output of the larger model, and for the ground truth labels by a weighted loss function.

\subsubsection{Why Distillation Can Introduce Bias:}
When computing distillation loss, we do not use the ground truth labels of the data points. Instead, we use the output of a model as labels. This can be problematic because the data corresponding to new classes can be more similar to some older classes than others. For example, if in an animal classifier one of the class is a whale, and others are terrestrial  animals, and the new classes are also all of terrestrial  animals, the older model would most probably assign a near zero probability of whale to the new images. This would introduce a bias against the whale class and result in poor performance of TC on whales.  This problem of bias was noticed by G.Hinton \etal \cite{hinton2015distilling} in their original paper as well. However, instead of proposing a strategy to remove this bias in a practical setting, they were only interested in showing the existence of a vector that can be used to remove the bias. As a result, they found a scaling vector by doing a grid search over the \textit{test} set. It's understood that we can not do such a search on the test set in a practical setting.

\subsection{Threshold Moving} 
Threshold moving is a well-known technique to tackle the issue of class imbalance. Buda~\etal~\cite{thresh} showed that threshold moving is effective in removing bias in ANNs. However, a measure of imbalance is required to apply the technique. In simple cases, the degree of imbalance can simply be measured by the frequency of instances of each class in the training set. In case of distillation, however, it is not clear how we can compute such a scaling vector because some classes might have zero frequency. Inspired by the loss function used for distillation, we propose that by simply measuring how much each class contributes to the target, it's possible to compute the scale vector.

\begin{figure}
	\centering
	\begin{tabular}{cc}
		
		\includegraphics[width=6.0cm]{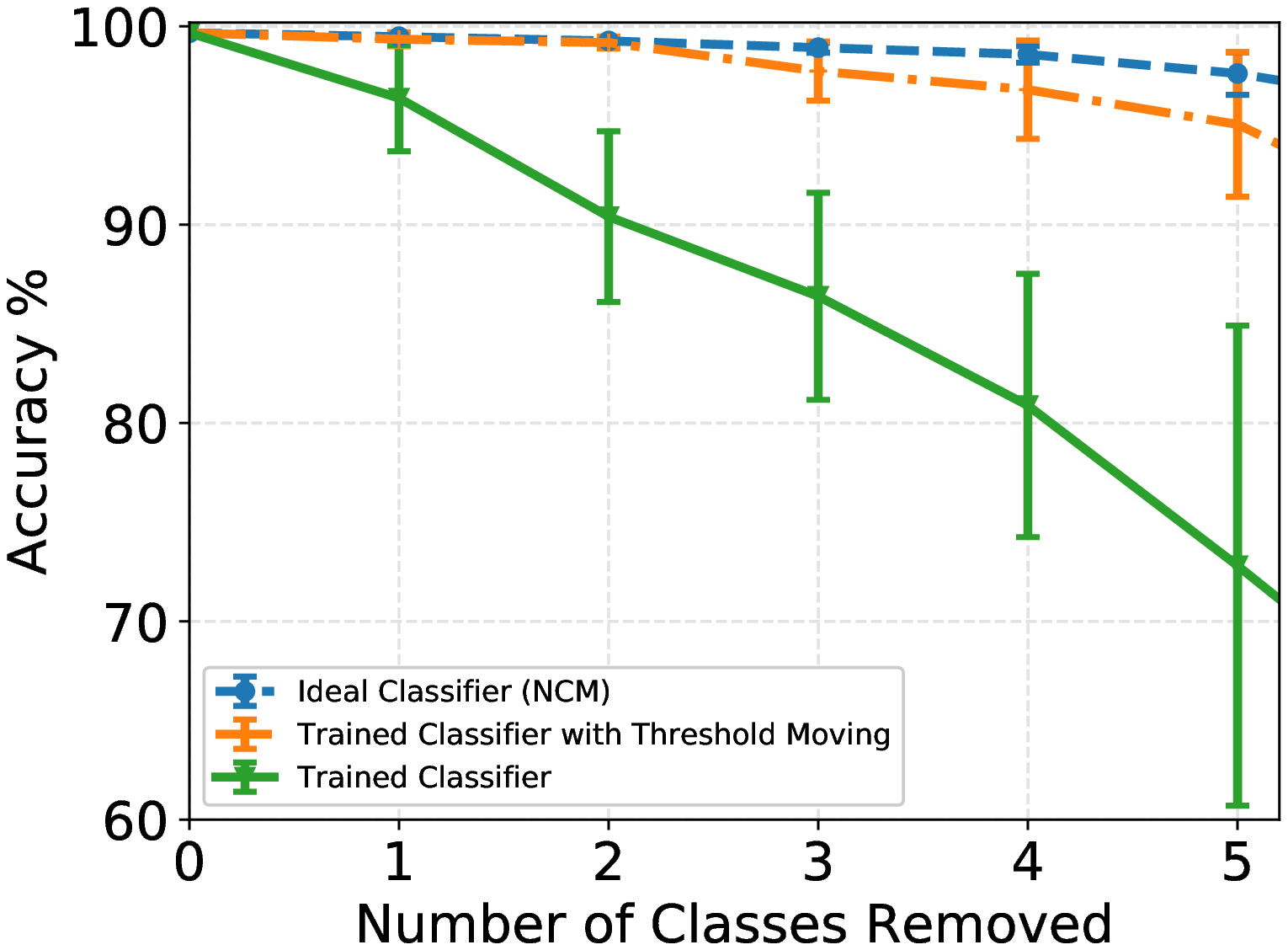}&
		\includegraphics[width=6.0cm]{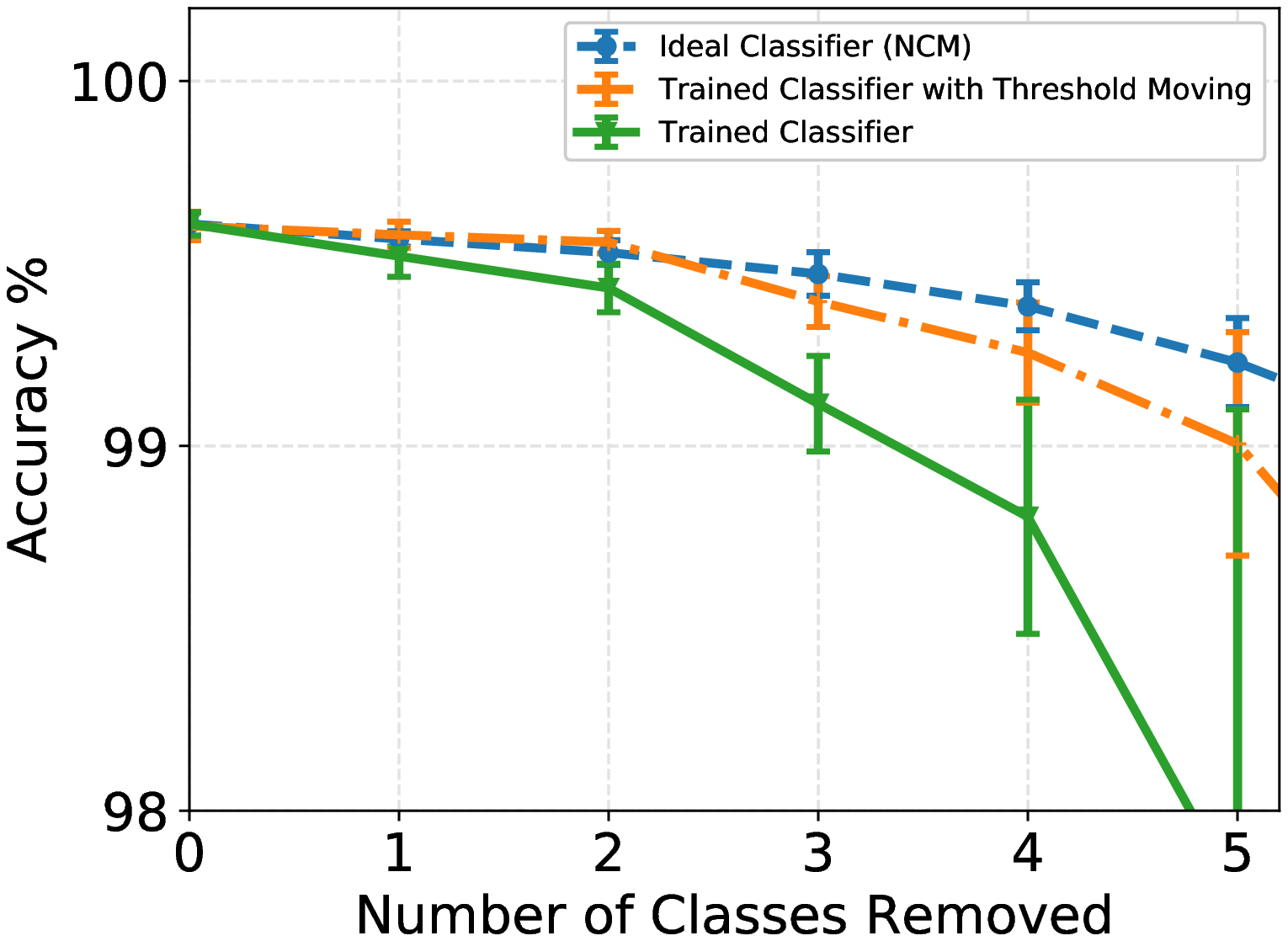}\\
		(a) Temperature = 2 &(b) Temperature = 5
	\end{tabular}
	\caption{Result of threshold moving with T = 2 and 5 on the unmodified test set of MNIST. We train our model after removing all instances of a certain no of classes (x-axis) from the train set, and only distill the knowledge of these classes from a teacher network trained on all data. Note that different scale is used for the y axis for a and b because using higher temperature results in softened targets and consequently, less bias.}
	\label{fig:dynamic}
\end{figure}

\subsection{Algorithm for Scale Computation:}
Let $F(x_i)$ be the model that outputs the probability distribution over $N$ classes i.e, $\forall x_i \in X, \; F(x_i) = P(n| x_i)$  where $ 0<=n<N $.  Suppose now that we want to find another model, $G(X)$, that gives a distribution over the old $N$ classes and $k$ new classes. Furthermore, we want to find $G$ given only the data of the $k$ new classes, and the original model $F(X)$. Finally, let $y_i $ be ground truth of new classes and $\mathcal D$ be the training dataset.
Z. Li and D. Hoiem \cite{li2017learning} showed that we can train such a model minimizing the following loss function:
\begin{equation}
\label{eq:hinton}
\sum_  {x_i, y_i \in \mathcal D}(1-\gamma)\times C_{entropy}(G(x_i), y_i) + T^2\gamma \times C_{entropy}(G^T(x_i), F^T(x_i))
\end{equation}
where $C_{entropy}$ is the cross-entropy loss function, $\mathcal D$ is the training dataset and $G^T(x_i)$ is exactly the same as $G(x_i)$ except we scale the values of the final logit of the neural network by $\frac{1}{T}$ before applying the final softmax . Note that we multiply the distillation loss by $T^2$ as suggested by G. Hinton \etal \cite{hinton2015distilling} to keep the magnitude of gradients equal as we change the temperature T.

This loss function, however, results in a biased classifier as discussed above. We demonstrate that scaling the predictions of $G(X)$ by a scale factor $\mathcal S$ given by the following equation is effective for bias removal.
\begin{equation}
\label{eq:scaling}
\mathcal S =  \sum_  {x_i, y_i \in \mathcal D}(1-\gamma) \times y_i + T^2\gamma \times F^T(x_i) 
\end{equation}
Note that Equation. \ref{eq:scaling} for scale vector computation is very similar to the distillation loss described in Equation. \ref{eq:hinton}. In fact, the scale vector is simply the sum of target probability distributions in the cross entropy loss. 
\\ \\
\begin{figure}
	\centering
	\begin{tabular}{cc}
		
		\includegraphics[width=5.5cm]{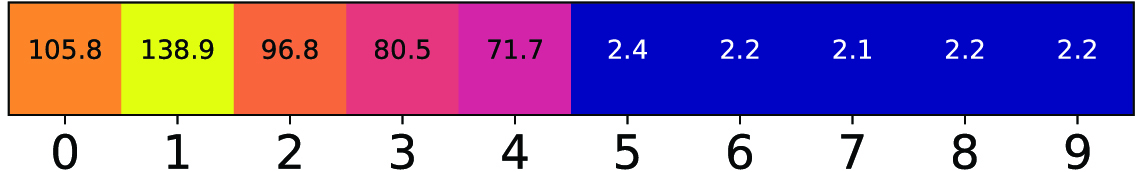}&
		\includegraphics[width=5.5cm]{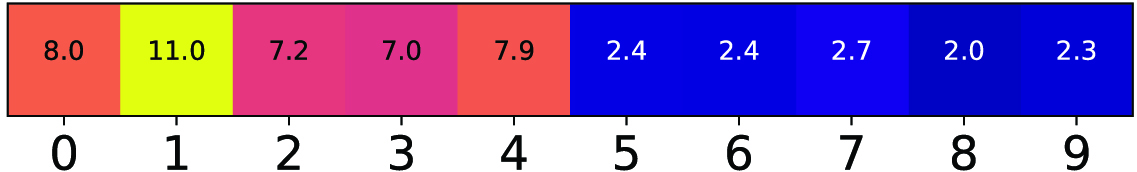}\\
		(a) Temperature = 2 &(b) Temperature = 5
	\end{tabular}
	\caption{Visualization of $\frac{\| \mathcal  S \|}{\mathcal  S}$ when the model is trained after removing first five MNIST classes using distillation. Note that the scaling factor is considerably larger for T=2 than T=5. This is because as we increase the temperature, targets are softened which reduces the bias.}
	\label{fig:scale}
\end{figure}

Our final prediction $G'(X)$ is then given by:
\begin{equation}
\label{eq:scaling1}
G'(X) = G(x)\circ \frac{\| \mathcal  S \|}{\mathcal  S}
\end{equation}

Where $\circ$ represents point-wise multiplication. Note that due the similarity between Equation~\ref{eq:hinton} and Equation~\ref{eq:scaling}, it is possible to compute $ \mathcal S$ during training at no additional cost. A visualization of the scaling factor $ \frac{\| \mathcal  S \|}{\mathcal  S}$ can be seen in Fig. \ref{fig:scale}.

\subsection{Intuition Behind the Algorithm} An intuitive understanding of Equation~\ref{eq:scaling} is that we are measuring the relative representation of each class in the training targets. More concretely, we are computing the expected value of predictions made by a model trained using Equation. \ref{eq:hinton}. When we scale our class predictions by the reciprocal of the expected value, we are effectively normalizing the class predictions such that expected value of predicting each class is equal. This in turn removes the bias introduced during training. 
\begin{figure}
	\centering
	\includegraphics[width=  0.97 \linewidth]{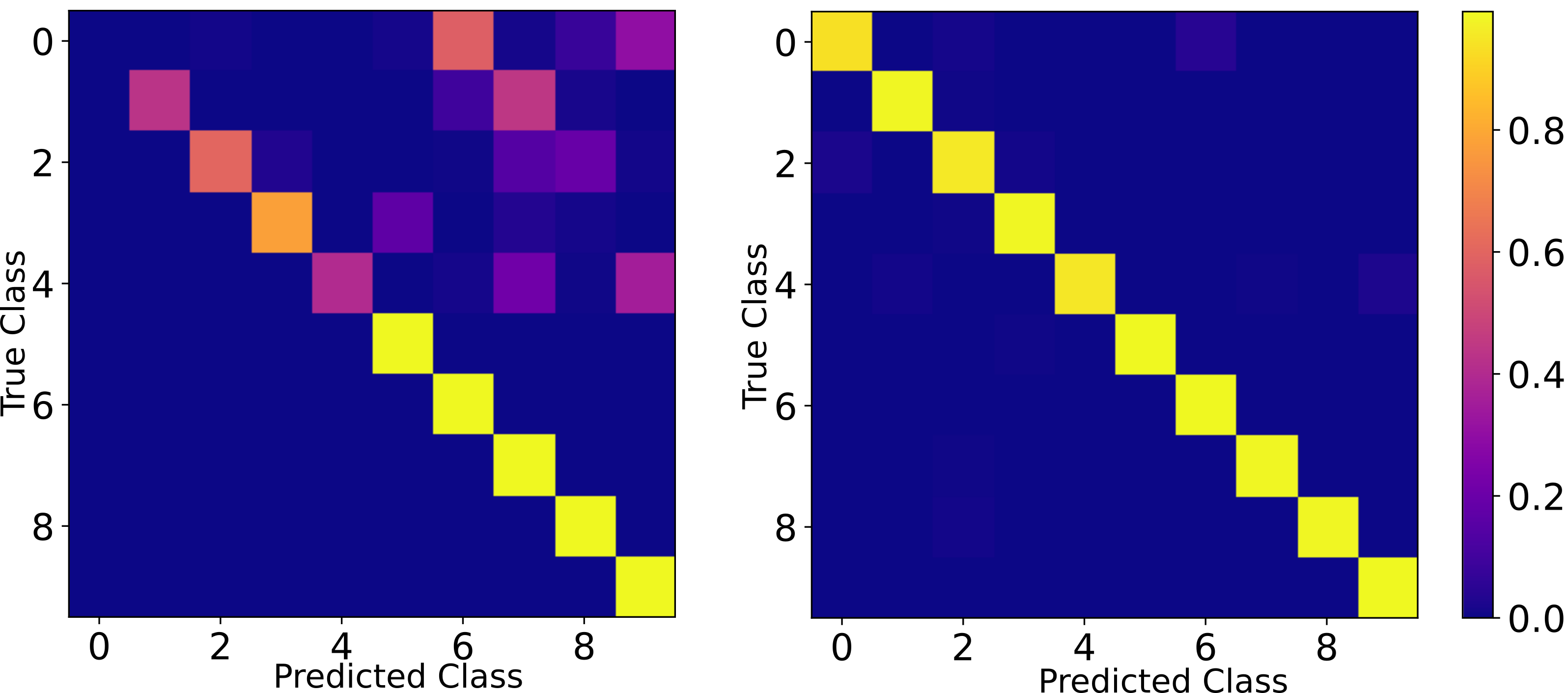}
	\caption{Confusion matrix of results of the classifier with (right) and without (left) threshold moving with T=2. We removed the first five classes of MNIST from the train set and only distilled the knowledge of these classes using a network trained on all classes. Without threshold moving the model struggled on the older classes. With threshold moving, however, not only was it able to classify unseen classes nearly perfectly, but also its performance did not deteriorate on new classes.}
	\label{confusion}
\end{figure}

\subsection{Experiment Design} To highlight and isolate the performance of our dynamic threshold moving algorithm, we first train a model on all the classes of MNIST dataset. We then train another randomly initialized model after removing $p$ randomly chosen classes from the MNIST dataset and use the first model to distill the information of the older classes. Note that the second model does not see any examples of the removed classes at train time. Finally, we test the second model on all classes of MNIST. We run this experiment for T equal two and five (The difference is visible for other values of T as well; we only choose two due to lack of space). 

\subsection{Results}
The results of the above-mentioned experiments can be seen in Fig.~\ref{fig:dynamic}. As evident from the figures, the accuracy of the Trained Classifier drops significantly as we remove more classes. However by simply scaling the predictions of the classifier, it is possible to recover most of the drop in the accuracy and achieve near optimal NCM classifier results (Note that NCM Classifier uses training data of all the classes at test time to find the mean embedding and as a result, is an ideal classifier).

A more detailed picture of the bias can be seen in Fig.~\ref{confusion}. The confusion matrix correspond to $x=5$ in Fig.~\ref{fig:dynamic} (a). It can be seen that without scaling, the model is struggling on the unseen classes. However after removing the bias, it can classify all 10 classes accurately. This is interesting because it shows that the model always had the discriminatory power to classify all classes and was performing poorly mainly because of the bias.

We further note that higher values of T results in smaller bias. Note that this is reflected in our computation of $\mathcal S$ in Equation. \ref{eq:scaling} where higher values of T lead to softened targets  $F^T(x_i)$ resulting in $\mathcal S$ with values closer to one. 

Finally, results of threshold moving on CIFAR100 are shown in  Fig.~\ref{fig:nmcvstc} (b). Again, we notice that our dynamic threshold algorithm is able to improve the performance of Trained Classifier to Trained Classifier Scaled. 
\section{Conclusion} 
In this paper, we analyzed the current state of the art method for class incremental learning, iCaRL, in detail. We showed that the primary reason iCaRL works well is not because it uses herding for instance selection, or Nearest Exemplar Mean classifier for classification, but rather because it uses the distillation loss on the exemplar set to retain the knowledge of older classes. We also proposed a dynamic threshold moving algorithm to fix the  problem of learned bias in the presence of distillation, and verified the effectiveness of the algorithm empirically. 

Finally, we release our implementation of an incremental learning framework implemented in a modern library that is easily extensible to new datasets and models, and allows for quick reproducibility of all of our results. We strongly believe that given the current state of research in the computer vision community, there is a strong need for analyzing existing work in detail, and making it easier for others to reproduce results and we hope that this work is a step in that direction. An anonymized version of our implementation is available at: \url{https://github.com/Khurramjaved96/incremental-learning}.

%
%
%
%
\bibliographystyle{splncs}
\bibliography{egbib}
\end{document}